\pdfoutput=1

\documentclass[11pt]{article}

\usepackage{EMNLP2022}

\usepackage{times}
\usepackage{latexsym}
\usepackage{makecell}
\usepackage{array}
\usepackage{multirow}
\usepackage[T1]{fontenc}

\usepackage[utf8]{inputenc}

\usepackage{microtype}

\usepackage{inconsolata}

\title{Summer: WeChat Neural Machine Translation Systems for the WMT22 Biomedical Translation Task}

\author{Ernan Li, Fandong Meng and Jie Zhou \\
Pattern Recognition Center, WeChat AI, Tencent Inc, China \\
\{cardli,fandongmeng,withtomzhou\}@tencent.com
}

\begin{document}

\maketitle

\begin{abstract}
This paper introduces WeChat's participation in WMT 2022 shared biomedical translation task on Chinese$\to$English. Our systems are based on the Transformer\cite{transformer2017}, and use several different Transformer structures to improve the quality of translation. 
In our experiments, we employ data filtering, data generation, several variants of Transformer, fine-tuning and model ensemble.
Our Chinese$\to$English system, named Summer, achieves the highest BLEU score among all submissions.
\end{abstract}

\section{Introduction}

This article describes the WeChat's participation in WMT 2022 shared biomedical translation task on Chinese$\to$English. We improve the translation quality of the system by increasing the diversity of model structure and data, fine-tuning the model with in-domain data, inserting tags at the beginning of each source sentence and selecting models with high diversity and good performance for ensemble.

For model architectures, our system adopt BIG and DEEP Transformer models which contain 10-layer and 20-layer encoders, 10240 and 4096 filter sizes, respectively, with TRANSFORMER-BIG setting~\cite{transformer2017}. In order to increase the diversity of the model, we use structures such as Average Attention Transformer (AAN)~\cite{zhangetalaan2018} and Mixed-AAN Transformer architecture~\cite{wechat21} in the decoder part.

For data generation, we use back-translation~\cite{sennrichetalimproving2016}, knowledge distillation~\cite{kim-rush-2016-sequence}, and forward-translation~\cite{wechat21} to improve data quality. And we use some data augmentation methods to improve the model robustness, such as adding synthetic noise and dynamic top-p sampling~\cite{wechat21}. Furthermore, according to the different sources of the corpora, we add tags at the beginning of the source sentence to perform domain adaptation.

For fine-tuning, we use in-domain bilingual corpus to fine-tune models from the general domain to the biomedical domain, and use target denoising~\cite{meng2020wechat} to improve the diversity of models and  mitigate training-generation discrepancy.

For model ensemble, we use Self-BLEU~\cite{selfbleu2018texygen} to evaluate the similarity between models. We take the prediction of one model as the reference and use the prediction of the other model to calculate the BLEU score. The higher the Self-BLEU score, the lower the diversity of the models.

In the remainder of this paper, we start with presenting the data strategy in Section 2. Then we describe our system details in Section 3. Section 4 presents the experimental results. Finally, we conclude our work in Section 5.
 
\section{Data}

In this section, we introduce the details of bilingual and monolingual data used in this shared task.

\subsection{Bilingual Corpus}
Our baseline model is trained with out-of-domain (OOD) data from WMT 2022 shared task on general machine translation\footnote{https://statmt.org/wmt22/translation-task.html}. Additionally, we use in-house data (depicted in Table~\ref{t:train_data} as OOD-IN-HOUSE) to improve performance of baseline model.
With regard to in-domain data, firstly, we use the in-domain bilingual corpus provided by the WMT 2022 shared biomedical translation task\footnote{https://github.com/biomedical-translation-corpora/corpora} (depicted in Table \ref{t:train_data} as IND-BIO). And we use the Champollion\footnote{http://champollion.sourceforge.net/} tool to align the sentences in the corpus. Then, we collect in-domain Chinese$\to$English (depicted in Table \ref{t:train_data} as IND-TAUS) sentence pairs from TAUS\footnote{https://taus-corona-corpus.s3.amazonaws.com/en-zh.txt.gz}.

\subsection{Monolingual Corpus}
The out-of-domain monolingual corpora are collected from WMT 2022 shared task on general machine translation and the in-house monolingual data.
With regard to in-domain data, the English part of the bilingual corpus in other languages provided by the WMT 2022 shared biomedical translation task is used as in-domain monolingual data.

\begin{table*}[t!]
\centering

\begin{tabular}{l c  c  c  c }
\hline
\bf \textsc{Language} & OOD-NEWS & OOD-IN-HOUSE & IND-BIO & IND-TAUS   \\
\hline
bilingual corpus  & 30.6M & 90M & 89K & 0.4M \\
monolingual corpus & 220M  & 50M & 6.9M & --  \\
\hline
\end{tabular}

\caption{Data used for training the system, where \emph{OOD-NEWS} is the out-of-domain data provided by WMT22 general translation task. \emph{OOD-IN-HOUSE} is the out-of-domain data collected from in-house corpus. \emph{IND-BIO} is the in-domain data provided by WMT22 shared biomedical translation task. And \emph{IND-TAUS} is the in-domain data collected  manually (not from MEDLINE, as depicted in 2.1). \emph{M} denotes \emph{million} and \emph{K} denotes \emph{thousand}.
} 
\label{t:train_data}
\end{table*}

\section{System overview}
In this section, we introduce the details of our system used in the WMT 2022 shared biomedical translation task. Our system adopts data filtering, data generation, model architectures, fine-tuning and ensemble.

\subsection{Data Filtering}
\label{sec:data_filter}
For data filtering, we use the following rules for bilingual corpus:
\begin{itemize}
\item Normalize punctuation with Moses scripts on both English and Chinese.
\item Filter out sentence pairs that are the same at the source and target.
\item Filter out sentence pairs whose source sentence's language recognition result is different from the original language.
\item Filter out sentence pairs with a source-to-target length ratio greater than 1:3.
\item Filter out the sentences longer than 150 words or exceed 40 characters in a single word.
\end{itemize}
Besides these rules, we use fast-align\footnote{https://github.com/clab/fast\_align} to filter out the sentence pairs with low alignment scores. We also filter out sentence pairs in which English sentences contain Chinese characters.

\subsection{Data Generation}
In this section, we introduce the approaches of data generation in our system, including back-translation, knowledge distillation, forward-translation, synthetic noise and tagging.

\subsubsection{Back-Translation}
Back-translation~\cite{hoang2018iterative} is the most commonly used data augmentation method in neural machine translation. Following the previous work~\cite{edunov-etal-2018-understanding}, we use following strategies to generate back translations to improve the diversity the training data:
\begin{itemize}
\item Beam search: We use beam search to generate the pseudo corpus with beam size setting to 4.
\item Dynamic top-p sampling: Following the work~\cite{wechat21}, at each decoding step, we select a word from the smallest set whose cumulative probability exceeds $p$, with $p$ varying from 0.9 to 0.95 during the data generation process.
\end{itemize}

\subsubsection{Knowledge Distillation}
For knowledge distillation~\cite{kim-rush-2016-sequence,wang2021selective}, we use the corpus generated from the teacher models to train the student models.

\subsubsection{Forward-Translation}
For forward-translation, we use an ensemble model to generate forward translations with the source-language monolingual corpus as input.

\subsubsection{Synthetic Noise}
For synthetic noise, we add different noises at the source side of the pseudo corpus to improve the diversity of the data and improve the robustness of the model:
\begin{itemize}
\item Randomly replace some source tokens with $<unk>$.
\item Randomly delete some tokens from the source sentence.
\item Randomly swap the two tokens in the source sentence in the specify window.
\end{itemize}

\subsubsection{Tagging}
For tagging, inspired by~\cite{johnson2017google}, we insert a tag at the beginning of each source sentence to denote its type: $<BT>$ for the back-translation data, $<NOISE>$ for the synthetic noise data, $<REAL>$ for the ground-truth bilingual corpus and $<FT>$ for the forward-translation data. Furthermore, we insert a tag at the second position of each sentence to denote its domain: $<BIO>$ for the in-domain data, $<NEWS>$ for the data from WMT22 general translation task and $<INHOUSE>$ for the data from our in-house corpus. At inference time, we always use the $<REAL>$ and $<BIO>$ tag. 

\subsection{Model Architectures}
In this section, we introduce the model architectures used by our system, including Transformer (Big/Deep), Average Attention Transformer (AAN) and Mixed Average Attention Transformer (Mixed-AAN)~\cite{wechat21}.

\subsubsection{Transformer}
Our baseline models are Big- and Deep-Transformer \cite{transformer2017} models. In our experiments, we use multiple model configurations with 20-layer and 30-layer encoders for deep models and 10-layers encoders for big models, and use 6-layers decoders for all models. The hidden size is set to 1024 and the filtering size is set from 4096 to 10240.

\subsubsection{Average Attention Transformer}
To increase the diversity between models, we adopt Average Attention Transformer \cite{zhangetalaan2018}, where the average attention is used to replace self-attention in the decoder. AAN summarizes the historical information of previous positions by means of cumulative average, which increases diversity with almost no harm to the quality of the model.

\subsubsection{Mixed-AAN Transformers}
Following the previous work~\cite{wechat21}, we adopt the Mixed-AAN Transformers to further improve the diversity and quality of models. In this experiment, we only use two architectures of Mixed-AAN:
\begin{itemize}
\item Self-first: In the decoder part, we use self-attention as the first layer, and then use average attention and self-attention alternately.
\item AAN-first: In the decoder part, we use average attention as the first layer, and then use self- attention and average attention alternately.
\end{itemize}

\subsection{Fine-tuning}
\label{sec:finetune}
For fine-tuning, we mainly use the in-domain data provided by WMT22 shared biomedical translation task for domain adaption~\cite{luong2015stanford, lietalniutrans2019}. In order to prevent the model from overfitting, as well as to improve the diversity of the model after domain transfer, we adopt target denoising \cite{meng2020wechat}. We add synthetic noise at the decoder inputs during fine-tuning. Therefore, with target denoising, the model becomes more robust. The method of adding synthetic noise is described in Section 3.2.4.

\subsection{Ensemble}
After obtaining a variety of different models through the above methods, we need to find the best model combination to get the best result. In general, the better the model performance and the greater the diversity between models, the better the performance for the model ensemble. To measure diversity, we use Self-BLEU~\cite{selfbleu2018texygen} to evaluate the similarity between models. Overall, we select 6 models from 52 candidate models for ensemble. All the candidate models are generated by different combinations of data and different training strategies as described earlier.

\begin{table*}[t!]
\begin{center}
\setlength{\tabcolsep}{6mm}{\begin{tabular}{l c c c c l l} 
\hline
\hline
\textbf{System}  & & & & \multicolumn{2}{c}{\textbf{BLEU}}  \\ 
\hline
\textbf{Baseline}  & & & & \multicolumn{2}{c} {34.57}    \\
$+$ IND-TAUS  & & & & \multicolumn{2}{c} {35.65}    \\
$+$ IND-BIO  & & & & \multicolumn{2}{c} {40.96}  \\
$+$ OOD-IN-HOUSE  & & & & \multicolumn{2}{c} {41.88}    \\
$+$ Back-Translation  & & & & \multicolumn{2}{c} {42.8}   \\
$+$ Knowledge Distillation  & & & & \multicolumn{2}{c} {43.12}    \\
$+$ Forward-Translation  & & & & \multicolumn{2}{c} {43.32}    \\
$+$ Multi BT  & & & & \multicolumn{2}{c} {44.11}   \\
\ \ \ \ \ $+$ Finetune  & & & & \multicolumn{2}{c} {44.96}  \\
\ \ \ \ \ $+$ Target denoise finetune  & & & & \multicolumn{2}{c} {45.1}  \\
\hline
\hline
\textbf{Baseline\_TAG}  & & & & \multicolumn{2}{c} {34.48}    \\
$+$ IND-TAUS  & & & & \multicolumn{2}{c} {35.62}    \\
$+$ IND-BIO  & & & & \multicolumn{2}{c} {41.07}  \\
$+$ OOD-IN-HOUSE  & & & & \multicolumn{2}{c} {42.14}    \\
$+$ Back-Translation  & & & & \multicolumn{2}{c} {43.91}   \\
$+$ Knowledge Distillation  & & & & \multicolumn{2}{c} {44.14}    \\
$+$ Forward-Translation  & & & & \multicolumn{2}{c} {44.39}    \\
$+$ Multi BT  & & & & \multicolumn{2}{c} {45.23}   \\
\ \ \ \ \ $+$ Finetune  & & & & \multicolumn{2}{c} {45.43}  \\
\ \ \ \ \ $+$ Target denoise finetune  & & & & \multicolumn{2}{c} {45.54}  \\
\hline
$+$ Ensemble & & & & \multicolumn{2}{c} {\bf 46.91$\star$}    \\
\hline
\hline
\end{tabular}
}
\end{center}

\caption{Translation performance on WMT21 biomedical translation task testset. $\star$ is the system we submitted. Multi BT means the iterative back-translation~\cite{hoang2018iterative} which use with different part of data and different generation strategies.}
\label{table:encn}
\end{table*}

\section{Experiments}
\subsection{Settings}
Our experiment is based on Fairseq~\footnote{https://github.com/pytorch/fairseq}.
The single models are carried out on 8 NVIDIA V100 / A100 GPUs.
We adopt the Adam optimizer with $\beta_{1}$ = 0.9, $\beta_{2}$ = 0.998. The batch-size is set to 4096 tokens, and the ``update-freq" is set to 4, and the warmup step is set to 4000 and the learning rate is set to 0.0005.

\subsection{Pre-processing and Post-processing}
The Chinese sentences are segmented by a in-house segmentation tool and English sentences are segmented by the tokenizer toolkit in Moses\footnote{http://www.statmt.org/moses/}. We normalize punctuation using  Moses scripts on both English and Chinese. For handling uppercase and lowercase of the English letters, we add a special token at the beginning of a word to denote uppercase (\_UU\_) and title case (\_U\_). By this way to reduce the size of the word list and reduce the difficulty of model training. For instance, \emph{"We are together NOW."} $\to$ \emph{"\_U\_ we are together \_UU\_ now."}. We use BPE \cite{sennrichetal2016bpe} with 32K operations for all the languages.

With the regard of post-processing, we use \emph{detokenizer.perl} on the English translations provided in Moses.

\subsection{Results}
The experimental results of Chinese$\to$English on WMT21 OK-aligned biomedical test set are shown in Table~\ref{table:encn}.  

Compared with the baseline model (Baseline\_TAG), the in-domain bilingual data ($+$IND-BIO) provided by WMT22 shared biomedical translation task brings a huge improvement, with 6.5 point increase in BLEU score. After adding the in-house out-of-domain corpus ($+$OOD-IN-HOUSE), we further gain +1.1 BLEU. We further obtain +1.8 BLEU by applying back-translation ($+$Back-Translat), and +0.23 BLEU by using knowledge distillation ($+$Knowledge Distillation), and +0.25 BLEU by using forward-translation ($+$ Forward-Transla). After using iterative back-translation~\cite{hoang2018iterative} ($+$Multi BT) described in Table \ref{table:encn}, we further achieve improvement of +0.84 BLEU. 

Additionally, we can find that the model with TAG was similar to the model without TAG in early stage experiments. As the number of data categories and data domains increases, the model with tags gradually demonstrates its advantages. Our best single model ($+$Target denoise fine) achieves 45.54 BLEU score, and we finally achieve 46.91 BLEU score by model ensemble ($+$Ensemble).

\section{Conclusion}
We introduce WeChat's participation in WMT 2022 shared biomedical translation task on Chinese$\to$English. Our system is based on the Transformer~\cite{transformer2017}, and uses several different Transformer structures such as Average Attention and Mixed-AAN to improve the performance. We use several data augmentation methods such as iterative back-translation, knowledge distillation, forward-translation and synthetic noise. We use tags to assist the model in domain learning and use in-domain fine-tuning with target denoising to domain transfer. Finally a Self-BLEU based ensemble method is used for model ensemble. Overall, our system achieves 46.91 BLEU score on WMT21 OK-aligned biomedical test set, and we achieve the highest BLEU score among all submissions.

\bibliography{wmt22bio}
\bibliographystyle{acl_natbib}

\appendix

\end{document}